\documentclass[runningheads]{llncs}
\usepackage[T1]{fontenc}
\usepackage{graphicx,verbatim}
\usepackage{colortbl}
\usepackage[table]{xcolor}
\usepackage{xcolor}
\usepackage{hyperref}

\begin{document}
\title{When Brains Disagree: Biological Ambiguity Underlies
the Challenge of Amyloid PET Synthesis from
Structural MRI}
\titlerunning{Biological Ambiguity Limits Amyloid PET Synthesis from Structural MRI}
%
\author{Louise E. G. Baron\inst{1,2} \and
Ross Callaghan\inst{3} \and
David M. Cash\inst{4,5} \and Philip S. J. Weston\inst{4,5} \and Hojjat Azadbakht\inst{3} \and Hui Zhang\inst{1,6}}
\authorrunning{Baron et al.}

\index{Baron, Louise E. G.}
\index{Callaghan, Ross}
\index{Cash, David M.}
\index{Weston, Philip S. J.}
\index{Azadbakht, Hojjat}
\index{Zhang, Hui}
%
\institute{Hawkes Institute, University College London, UK \and
Department of Medical Physics and Biomedical Engineering, University College London, UK \and AINOSTICS Ltd, Manchester, UK \and Dementia Research Centre, UCL Queen Square Institute of Neurology, University College London, UK \and  UK Dementia Research Institute, London, UK \and
Department of Computer Science, University College London, UK 
\email{louise.baron.22@ucl.ac.uk}
}


\maketitle              
\begin{abstract}
Structural MRI-to-amyloid PET synthesis has been proposed as a non-invasive alternative for amyloid assessment in Alzheimer’s disease (AD). However, reported performance of identical models varies widely across studies, and increasingly complex architectures have not led to consistent gains. This inconsistency is thought to be caused by a fundamental biological ambiguity: MRI captures neurodegeneration, while PET measures amyloid pathology -- two processes that are often temporally decoupled in AD. As a result, similar MRI patterns may correspond to different amyloid states, creating ambiguous one-to-many mappings. MRI-to-amyloid PET synthesis may therefore be intrinsically ill-posed; however, this idea has yet to be tested scientifically.
The aim of this work is to test this hypothesis through two controlled experiments.
We first control the training distribution by stratifying paired MRI-PET data by amyloid and neurodegeneration status. Using two standard synthesis models under a controlled design, we show that biologically unambiguous mappings are learnable in isolation, but performance collapses when data ambiguity is introduced. This demonstrates that ambiguity in the data distribution, rather than architectural capacity, constrains performance.
Second, we show that introducing orthogonal biological information in the form of plasma biomarkers resolves this ambiguity. When multimodal inputs are incorporated, performance improves and stability is restored.
Together, these findings suggest that limited and inconsistent performance in MRI-to-amyloid PET synthesis is explained by intrinsic biological ambiguity, and that stable, meaningful progress requires multimodal integration rather than architectural complexity.

\end{abstract}

\section{Introduction}
Alzheimer’s disease (AD) is the leading cause of dementia, which affects over 55 million people worldwide and imposes a substantial and growing health and economic burden~\cite{who2021,Aranda2021}. After decades of symptom-based treatment, the therapeutic landscape of AD is beginning to change with the recent introduction of disease-modifying therapies that directly target amyloid-$\beta$ -- the key pathological hallmark of AD -- and offer the potential to slow disease progression~\cite{donanemab,lecanemab}. To access these therapies, patients must demonstrate significant amyloid burden, making accurate and scalable amyloid assessment central to modern AD care.

Amyloid positron emission tomography (PET) provides a direct, in vivo measure of amyloid pathology and is therefore a key tool for diagnosis and treatment eligibility~\cite{Kamatham2024,Chapleau2022}. However, amyloid PET is not routinely acquired due to its high costs, limited availability, and exposure to ionising radiation~\cite{VandeVrede2025}. By contrast, magnetic resonance imaging (MRI) is widely available and routinely acquired but primarily reflects downstream neurodegeneration rather than amyloid pathology. Relying on MRI alone therefore limits pathology-specific diagnosis and hinders access to amyloid-targeting therapies.

To bridge this gap, deep learning models have been proposed to synthesise amyloid PET from MRI~\cite{Vega2024,Jin2024,Ou2025,sargood2025cocolit,Chen2025_plasma,Chen2026,Theodorou2025}. Although a promising approach for non-invasive and scalable surrogates of amyloid imaging, existing work reports only moderate and highly inconsistent performance. Reported accuracies for classifying synthetic PET as amyloid positive/negative span a wide range -- from near chance~\cite{sargood2025cocolit,Chen2025_plasma} to over 80\%~\cite{Ou2025,Chen2026} -- even for the same baseline models and datasets. Moreover, many studies emphasise global image-quality metrics, whereas clinically meaningful amyloid-specific performance is inconsistently evaluated~\cite{Vega2024,Theodorou2025}. As a result, the true diagnostic value of MRI-derived amyloid PET remains unclear. This variability and inconsistency among studies suggest that limitations may arise not only from model design but also from the data itself, notably the biological relationship between the processes measured by MRI and amyloid PET. 

\begin{figure}
    \centering
    \includegraphics[width=3.2in]{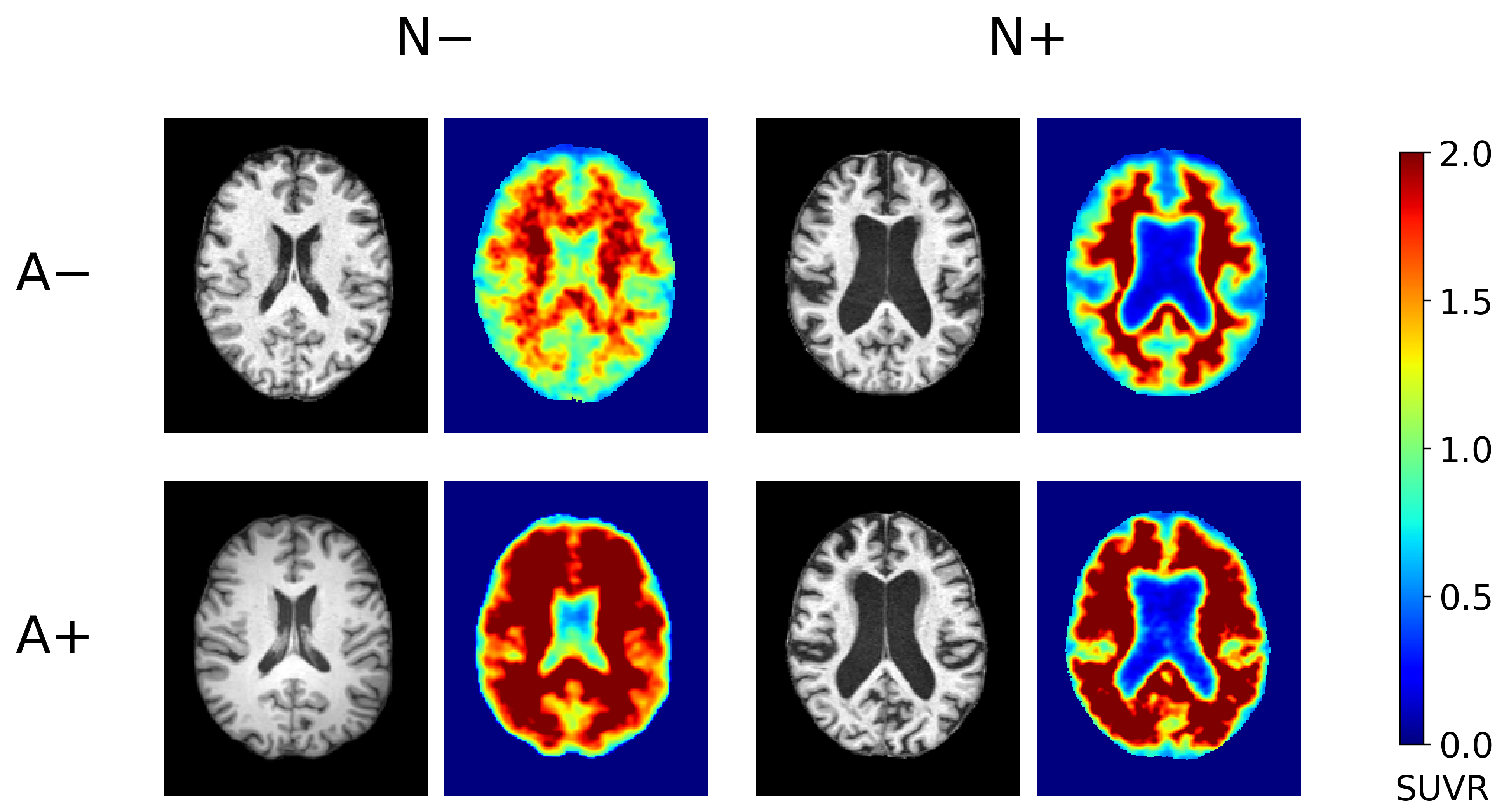}
    \caption{Distribution of amyloid (A) and neurodegeneration (N) profiles in ADNI subjects. Comparable patterns of structural neurodegeneration occur across different amyloid states, indicating that MRI-derived signals may not uniquely determine amyloid burden. Amyloid status was defined using a cortical SUVR threshold of 1.11~\cite{Landau2012}, and neurodegeneration was quantified using AVRA-derived atrophy scores~\cite{avra}.}
    \label{neuro_figs}
\end{figure}

According to the widely accepted model of AD progression~\cite{Jack2018}, amyloid accumulation can occur years to decades before measurable neurodegeneration, and structural decline may continue after amyloid burden has stabilised. As a result, individuals with similar MRI patterns may exhibit different amyloid loads, and vice versa. The coexistence of distinct biological states, amyloid positive or negative (A+/A--) with or without neurodegeneration (N+/N--), suggests a one-to-many mapping, where MRI may not uniquely determine amyloid PET.
Figure~\ref{neuro_figs} illustrates this phenomenon in subjects from the Alzheimer's Disease Neuroimaging Initiative (ADNI). 
If MRI does not uniquely determine amyloid pathology, then MRI-to-PET synthesis may be fundamentally ambiguous. In such a setting, increasing model complexity is unlikely to help, because the ambiguity arises from the data itself. 

Moreover, recent studies have shown that plasma biomarkers correlate with PET-derived measures of amyloid pathology~\cite{plasma1,plasma2,plasma3}. Although they do not directly image the brain, they provide an independent molecular signal that is not encoded in structural anatomy. As such, they offer a principled way to examine whether instability in MRI-based synthesis reflects missing biological information rather than architectural challenges. 

In our study, we hypothesise that biological decoupling between amyloid and neurodegeneration makes MRI-to-amyloid PET synthesis an ill-posed problem, where MRI does not uniquely determine amyloid PET. We test this hypothesis through two controlled experiments that examine the effect of biological ambiguity and the role of additional molecular information on mapping stability.

\section{Experimental Design}
To test our hypothesis, we propose the following experiments:

\textbf{(i) Ambiguity isolation:} We manipulate the composition of the training data to explicitly control ambiguity. If ambiguity is the limiting factor, then different biological profiles should be learnable in isolation, while performance should deteriorate when conflicting profiles are mixed.

\textbf{(ii) Biological completion:} If degradation and instability arise because MRI alone is insufficient to uniquely predict amyloid, then introducing complementary molecular signals should reduce ambiguity and improve stability. We therefore test whether conditioning on plasma biomarkers stabilises the mapping.


\subsection{Experiment I: Isolating Biological Ambiguity}

To isolate the effect of biological ambiguity, we trained models under three data regimes: a
\textit{Baseline} model trained on all available data (all A/N pairs), a
\textit{Concordant} model trained only on biologically aligned cases (A--/N-- and A+/N+), and a
\textit{Discordant} model trained only on biologically dissociated cases (A--/N+ and A+/N--). 
After training, all models were evaluated on the same fixed held-out test set, which included all A/N combinations.

\textbf{Dataset and biological A/N definition} were based on 2{,}884 paired T1w MRI and amyloid PET scans (AV45 tracer) from 1{,}358 subjects in ADNI. The cohort was sex-balanced (51\% female), with a mean age of 74 years, and contributed between 1 and 7 scan pairs per subject (mean 2.1). The average time interval between MRI and PET within each pair was 19 days. 
\textit{Amyloid status (A--/A+)} was assigned using the standard cortical SUVR threshold of 1.11 defined by ADNI~\cite{Landau2012}. 
\textit{Neurodegeneration status (N--/N+)} was quantified using AVRA~\cite{avra}, an automated visual rating framework that estimates established structural atrophy scores from T1w MRI. Since our goal was to capture global neurodegeneration signals rather than region-specific atrophy patterns, we summarised AVRA outputs into a single neurodegeneration score by averaging all AVRA scores. 
To obtain clearly separable groups, we defined N-- as AVRA $\leq 0.35$ and N+ as AVRA $\geq 0.8$. These cut-offs were chosen via visual inspection of the AVRA distribution and corresponding structural patterns to capture clearly distinct atrophy states, and our conclusions are not expected to depend strongly on small variations of these. Subjects with $0.35 < \mathrm{AVRA} < 0.8$ were labelled as N-\emph{intermediate}.

\textbf{Two 3D image synthesis architectures} were evaluated: the widely benchmarked conditional GAN pix2pix~\cite{Isola2017} and a latent diffusion model (LDM)~\cite{Rombach2021,Zhang2023}, which has recently been applied to amyloid PET synthesis \cite{Ou2025,sargood2025cocolit}. All models were trained using identical data splits and preprocessing.

\textbf{Train-test splitting} was performed at the subject level to avoid leakage across splits, and the training distribution was balanced across A/N groups to ensure comparable representation of biological profiles. A total of 1,164 image pairs were used for training (with 10\% for validation), and 1,720 pairs for testing. The training sample size is comparable to prior studies~\cite{Vega2024,Ou2025,Chen2025_plasma,sargood2025cocolit}, while the much larger test set enables more stable and statistically reliable performance estimates than typically reported in the literature (100-300 subjects).

\textbf{Image preprocessing} involved co-registering MRI and PET and spatially aligning them to the MNI-152 template using affine transformations. MRI images were normalised ($0.1$st and $99.9$th percentiles) to $[0,1]$. PET volumes were kept in SUVR space to preserve quantitative amyloid information.

\textbf{Evaluation} of the models was performed using both global image quality metrics and amyloid-specific clinical metrics. Image quality was assessed using peak signal-to-noise ratio (PSNR) and structural similarity index measure (SSIM). Clinical validity was assessed by measuring reconstruction accuracy of cortical amyloid burden and by evaluating A--/A+ classification performance. 
Performance is reported on the full test set and in each of the previously defined biological subsets.

\subsection{Experiment II: Biological Completion}

Experiment I tested whether instability emerges when biologically conflicting states are mixed. To examine whether this instability reflects incomplete biological information in MRI alone, we conditioned the Baseline model on plasma biomarkers and evaluated it under the same experimental framework.

\textbf{Plasma data} from ADNI were available for only a subset of the dataset from Experiment I. In total, 589 MRI-PET pairs had corresponding plasma measurements.
We incorporated markers that provide molecular information related to amyloid burden and disease progression not directly observable in structural MRI, specifically phosphorylated tau 217 (pT217), amyloid-$\beta$42 (A$\beta$42), amyloid-$\beta$40 (A$\beta$40), neurofilament light chain (NfL), and glial fibrillary acidic protein (GFAP).

\textbf{Plasma-conditioned synthesis} was implemented via feature-wise linear modulation (FiLM)~\cite{perez2018film} in the pix2pix U-Net generator. A small multilayer perceptron projected the plasma vector to scaling and shifting parameters that modulated the U-Net bottleneck and conditioned the MRI-to-amyloid PET mapping on complementary molecular information. In this experiment, architecture was kept fixed to focus on the effect of biological completion.

\textbf{Train-test split and preprocessing} followed the same approach as in Experiment I. Out of the 589 MRI-PET pairs with plasma data, 390 were used for training and 199 for testing. To allow direct comparison between the unimodal and plasma-conditioned model, plasma cases retained their original train/test assignment. Plasma values were also z-score normalised using training-set statistics.

\textbf{Evaluation} of the plasma-conditioned model was performed using the same metrics as in Experiment I. For direct comparison, performance was evaluated against the MRI-only Baseline on the same test set (n = 199).

\section{Results and Discussion}

\subsection{Experiment I: Isolating Biological Ambiguity}

\begin{figure}[ht]
    \centering
    \includegraphics[width=\textwidth]{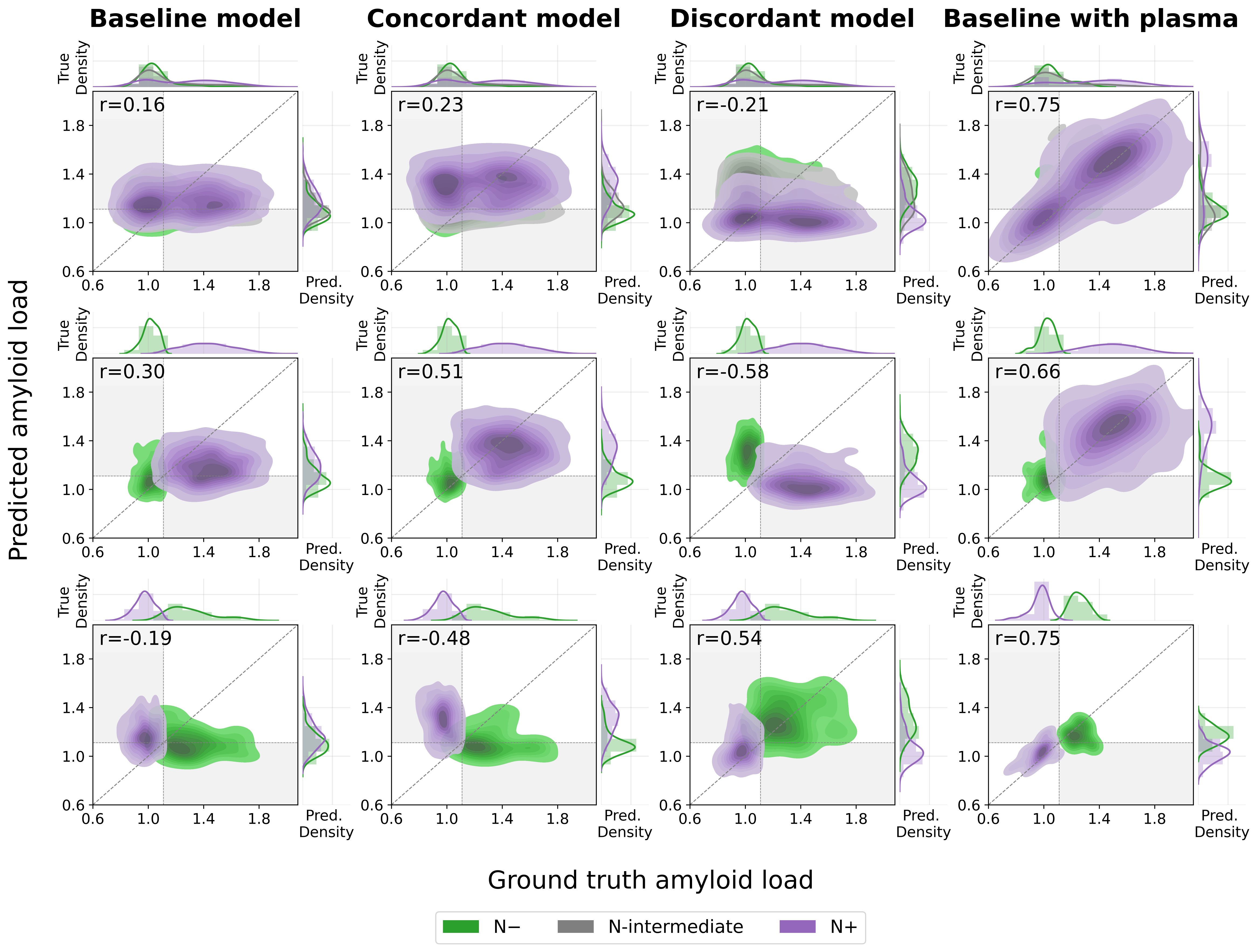}
    \caption{Amyloid load prediction across the three training regimes from Experiment I (Baseline, Concordant, and Discordant) and the plasma-conditioned model from Experiment II, using the pix2pix architecture. Columns correspond to model variants, and rows show evaluation on the \textbf{full test set (top)}, \textbf{concordant subset (middle)}, and \textbf{discordant subset (bottom)}.}
    \label{pix2pix-plots}
\end{figure}

Performance of the three (Baseline, Concordant, Discordant) models across different test sets is summarised in Table~\ref{pix2pix-ldm-table-by-model}, with prediction distributions for the pix2pix model (true vs predicted amyloid load) shown in Figure~\ref{pix2pix-plots}.




\begin{table}[ht]
\centering
\caption{Performance across three training regimes organised by architecture, training regime, and test set. Values are mean $\pm$ standard deviation (B=500 bootstrap). For each model, best performance across test sets is shown in bold. B-ACC: Balanced Accuracy, SEN: Sensitivity, SPEC: Specificity, CORR: Pearson correlation of mean cortical amyloid load.}
\label{pix2pix-ldm-table-by-model}

{\fontsize{8}{9.5}\selectfont
\renewcommand{\arraystretch}{1.1}

\begin{tabular}{l c c c c c c}
\hline
\textbf{Test} &
\textbf{PSNR} &
\textbf{SSIM} &
\textbf{B-ACC(\%)} &
\textbf{SEN(\%)} &
\textbf{SPEC(\%)} &
\textbf{CORR} \\
\hline

\rowcolor{gray!12}
\multicolumn{7}{l}{\textbf{3D pix2pix}} \\
\hline

\multicolumn{7}{l}{\textit{\textbf{Baseline model}}} \\[-1pt]
All test
& \bf{21.2} $\pm$ 0.1
& 0.905 $\pm$ 0.001
& 57.4 $\pm$ 1.7
& 66.5 $\pm$ 2.5
& 48.2 $\pm$ 2.7
& 0.16 $\pm$ 0.04 \\
Conc. test
& 20.8 $\pm$ 0.1
& 0.902 $\pm$ 0.001
& \bf{69.0} $\pm$ 2.7
& \bf{74.5} $\pm$ 2.5
& \bf{63.6} $\pm$ 4.8
& \bf{0.30} $\pm$ 0.05 \\
Disc. test
& 21.0 $\pm$ 0.2
& \bf{0.911} $\pm$ 0.002
& 36.0 $\pm$ 4.8
& 43.0 $\pm$ 8.8
& 29.0 $\pm$ 4.1
& -0.19 $\pm$ 0.06 \\
\hline

\multicolumn{7}{l}{\textit{\textbf{Concordant model}}} \\[-1pt]
All test
& 21.0 $\pm$ 0.1
& 0.902 $\pm$ 0.001
& 58.2 $\pm$ 1.4
& 80.5 $\pm$ 1.8
& 35.9 $\pm$ 2.5
& 0.23 $\pm$ 0.03 \\
Conc. test
& \bf{21.2} $\pm$ 0.1
& 0.901 $\pm$ 0.001
& \bf{82.3} $\pm$ 2.4
& \bf{95.0} $\pm$ 1.2
& \bf{69.7} $\pm$ 4.6
& \bf{0.51} $\pm$ 0.04 \\
Disc. test
& 19.8 $\pm$ 0.3
& \bf{0.904} $\pm$ 0.002
& 19.0 $\pm$ 4.1
& 29.5 $\pm$ 7.8
& 8.4 $\pm$ 2.4
& -0.48 $\pm$ 0.06 \\
\hline

\multicolumn{7}{l}{\textit{\textbf{Discordant model}}} \\[-1pt]
All test
& 20.4 $\pm$ 0.1
& 0.900 $\pm$ 0.001
& 40.1 $\pm$ 1.7
& 45.4 $\pm$ 2.3
& 34.8 $\pm$ 2.7
& -0.21 $\pm$ 0.04 \\
Conc. test
& 19.3 $\pm$ 0.1
& 0.893 $\pm$ 0.001
& 15.3 $\pm$ 1.4
& 26.0 $\pm$ 2.2
& 4.7 $\pm$ 1.3
& -0.58 $\pm$ 0.03 \\
Disc. test
& \bf{21.7} $\pm$ 0.2
& \bf{0.912} $\pm$ 0.002
& \bf{83.2} $\pm$ 2.7
& \bf{94.4} $\pm$ 3.8
& \bf{72.0} $\pm$ 4.0
& \bf{0.54} $\pm$ 0.07 \\
\hline

\rowcolor{gray!12}
\multicolumn{7}{l}{\textbf{3D LDM }} \\
\hline

\multicolumn{7}{l}{\textit{\textbf{Baseline model}}} \\[-1pt]
All test
& 18.3 $\pm$ 0.1
& 0.824 $\pm$ 0.001
& 50.8 $\pm$ 1.4
& 39.7 $\pm$ 1.9
& 62.0 $\pm$ 1.9
& 0.05 $\pm$ 0.03 \\
Conc. test
& 17.9 $\pm$ 0.1
& 0.822 $\pm$ 0.001
& 47.5 $\pm$ 2.1
& 37.8 $\pm$ 2.4
& 57.2 $\pm$ 3.5
& 0.05 $\pm$ 0.05 \\
Disc. test
& \bf{18.6} $\pm$ 0.1
& \bf{0.830} $\pm$ 0.002
& \bf{64.2} $\pm$ 4.1
& \bf{62.4} $\pm$ 7.5
& \bf{65.9} $\pm$ 3.3
& \bf{0.12} $\pm$ 0.06 \\
\hline
\multicolumn{7}{l}{\textit{\textbf{Concordant model}}} \\[-1pt]
All test
& 20.0 $\pm$ 0.1
& 0.877 $\pm$ 0.001
& 61.4 $\pm$ 1.6
& 70.7 $\pm$ 2.2
& 52.2 $\pm$ 2.5
& 0.26 $\pm$ 0.04 \\
Conc. test
& \bf{20.4} $\pm$ 0.1
& \bf{0.877} $\pm$ 0.001
& \bf{85.8} $\pm$ 1.7
& \bf{88.1} $\pm$ 1.8
& \bf{83.5} $\pm$ 3.0
& \bf{0.58} $\pm$ 0.03 \\
Disc. test
& 18.4 $\pm$ 0.2
& 0.874 $\pm$ 0.002
& 18.7 $\pm$ 3.9
& 25.9 $\pm$ 7.4
& 11.5 $\pm$ 2.5
& -0.45 $\pm$ 0.06 \\
\hline

\multicolumn{7}{l}{\textit{\textbf{Discordant model} }} \\[-1pt]
All test
& 19.5 $\pm$ 0.1
& 0.875 $\pm$ 0.001
& 39.8 $\pm$ 1.5
& 28.3 $\pm$ 2.0
& 51.3 $\pm$ 2.4
& -0.19 $\pm$ 0.03 \\
Conc. test
& 18.4 $\pm$ 0.1
& 0.867 $\pm$ 0.001
& 15.7 $\pm$ 1.9
& 12.3 $\pm$ 1.6
& 19.1 $\pm$ 3.5
& -0.53 $\pm$ 0.03 \\
Disc. test
& \bf{21.0} $\pm$ 0.2
& \bf{0.888} $\pm$ 0.002
& \bf{80.9} $\pm$ 3.0
& \bf{79.1} $\pm$ 5.3
& \bf{82.7} $\pm$ 3.1
& \bf{0.57} $\pm$ 0.04 \\
\hline

\end{tabular}
}
\end{table}

\textbf{Simplified biological regimes yield learnable mappings.}
The Concordant and Discordant models were designed as controlled experiments to reduce biological ambiguity in the MRI-to-amyloid relationship. Within their respective in-distribution evaluations, both models achieved strong amyloid-specific performance, with balanced accuracies of 82.3\% (corr = 0.51) and 83.2\% (corr = 0.54), respectively with the pix2pix model. These results demonstrate that the synthesis architecture \textit{can} learn amyloid mappings when the data is biologically consistent, establishing a best-case scenario in which the task is learnable.

\textbf{Mixing biological regimes degrades the learned mapping.}
As seen in the Baseline model performance, when the same architecture was trained on the full dataset containing all biological states, performance decreased substantially (57.4\% balanced accuracy, corr = 0.16), despite exposure to a larger and more diverse training set. Notably, the Concordant model, trained on fewer and less diverse samples, achieved comparable performance when evaluated across the full test distribution (58.2\% balanced accuracy). This demonstrates that increasing data quantity alone does not improve performance when the training data contains incompatible biological relationships. 

\textbf{Performance degradation reflects learning shortcuts under conflicting supervision.}
When faced with biologically conflicting signals, the Baseline model showed signs of converging toward an averaged solution, including unstable performance across biological subsets, strong imbalance between sensitivity and specificity, and compression of the predicted amyloid dynamic range relative to ground truth (Figure~\ref{pix2pix-plots}). Together, these findings suggest that mixing conflicting biological mappings encourages learning shortcuts, where the model favours an average prediction rather than learning biologically grounded mappings.

\textbf{The same ambiguity-dependent pattern is observed with latent diffusion.} Trained under the same controlled regimes, the LDM exhibited the same behaviour as pix2pix: strong in-distribution performance for biologically consistent training (Concordant and Discordant models) but significant degradation when conflicting biological states were mixed in training (Baseline model). Although absolute image-quality metrics differed between architectures, the pattern of performance across biological subsets was preserved. This indicates that the instability arises from the data distribution rather than generative architecture. 

\subsection{Experiment II: Biological Completion}

\textbf{Multimodal conditioning provides additional evidence of an ambiguous mapping.}
As shown in Table~\ref{tab:plasma_comparison}, adding plasma as a conditioning variable improved balanced accuracy from 64.5\% (MRI-only Baseline) to 85.3\% (MRI+plasma) when both models were evaluated on the same plasma-available test subset (n = 199). This improvement also came with the recovery of the amyloid load dynamic range, with an increase in correlation from 0.21 to 0.75 (Figure~\ref{pix2pix-plots}), and more balanced sensitivity and specificity, suggesting that the model no longer relied on an averaged shortcut solution. Importantly, performance remained consistent between the concordant and discordant subsets, suggesting that additional molecular information helps disentangle the signal from the MRI input. Together, these results provide independent evidence that the degradation observed in the MRI-only Baseline model reflects a biologically underdetermined mapping rather than limitations related to model architecture or data abundance.

\begin{table}[t]
\centering
\caption{Performance of the MRI-only and MRI+plasma-conditioned Baseline model on the plasma test set (n = 199). Values are mean $\pm$ standard deviation (B=500 bootstrap). B-ACC: Balanced Accuracy, SEN: Sensitivity, SPEC: Specificity, CORR: Pearson correlation of mean cortical amyloid load.}
\label{tab:plasma_comparison}

{\fontsize{8}{9.5}\selectfont
\setlength{\tabcolsep}{2.2pt}
\renewcommand{\arraystretch}{1.1}

\begin{tabular}{l c c c c c c}
\hline
\textbf{Test} &
\textbf{PSNR} &
\textbf{SSIM} &
\textbf{B-ACC(\%)} &
\textbf{SEN(\%)} &
\textbf{SPEC(\%)} &
\textbf{CORR} \\
\hline

\multicolumn{7}{l}{\textbf{Baseline (MRI Only)}} \\[2pt]

All test        
& 20.8 $\pm$ 0.2
& 0.907 $\pm$ 0.002
& 64.5 $\pm$ 3.8 
& 71.6 $\pm$ 5.2 
& 57.4 $\pm$ 5.6 
& 0.21 $\pm$ 0.07 \\

Conc. test 
& 20.1 $\pm$ 0.3
& 0.902 $\pm$ 0.003
& 75.7 $\pm$ 4.8 
& 77.4 $\pm$ 5.6 
& 74.1 $\pm$ 8.0 
& 0.31 $\pm$ 0.09 \\

Disc. test 
& 21.7 $\pm$ 0.5
& 0.914 $\pm$ 0.003
& 23.3 $\pm$ 8.3 
& 12.7 $\pm$ 12.4 
& 33.8 $\pm$ 10.8 
& -0.50 $\pm$ 0.14 \\

\hline

\multicolumn{7}{l}{\textbf{Baseline (MRI + Plasma)}} \\[2pt]

All test        
& 21.5 $\pm$ 0.2
& 0.910 $\pm$ 0.003
& 85.3 $\pm$ 2.6 
& 88.7 $\pm$ 3.2 
& 82.1 $\pm$ 4.3 
& 0.75 $\pm$ 0.06 \\

Conc. test 
& 21.1 $\pm$ 0.3
& 0.906 $\pm$ 0.004
& 84.2 $\pm$ 4.3 
& 94.1 $\pm$ 2.9 
& 74.4 $\pm$ 8.1 
& 0.66 $\pm$ 0.09 \\

Disc. test 
& 22.4 $\pm$ 0.5
& 0.917 $\pm$ 0.003
& 82.5 $\pm$ 9.2 
& 73.9 $\pm$ 17.3 
& 91.1 $\pm$ 6.1 
& 0.75 $\pm$ 0.09 \\

\hline
\end{tabular}}
\end{table}

\section{Conclusion}

Prior work on MRI-to-amyloid PET synthesis reports widely varying and inconsistent performance despite similar architectures and datasets. Our findings suggest the following explanation: MRI does not uniquely determine amyloid PET. Because neurodegeneration and amyloid pathology are temporally decoupled processes in AD, the MRI-to-amyloid PET mapping is biologically ambiguous.
We have shown that under this biological ambiguity, performance is unstable across architectures, and accurate mappings are only learnable when ambiguity is explicitly controlled. Moreover, introducing complementary molecular information through plasma biomarkers restores stable performance, indicating that the limitation arises from missing biological signal rather than model architecture.
Together, these results suggest that instability in MRI-to-amyloid PET synthesis reflects an ill-posed problem. Meaningful progress will therefore require explicitly accounting for biological heterogeneity and shifting towards multimodal solutions that complement the MRI signal, rather than focusing solely on increasingly complex architectures. Future work with larger multimodal cohorts will allow further validation of these findings and assess the generalisability of biological conditioning across architectures.


\begin{credits}
\subsubsection{\ackname} L. E. G. Baron is supported by the EPSRC-funded UCL Centre for Doctoral Training in Intelligent, Integrated Imaging in Healthcare (i4health) [EP/S021930/1]. P. S. J. Weston is supported by the Wellcome Trust, NIHR, and NIH.
Data used in preparation of this article were obtained from the Alzheimer's Disease Neuroimaging Initiative (ADNI) database (\url{adni.loni.usc.edu}). As such, the investigators within the ADNI contributed to the design and implementation of ADNI and/or provided data but did not participate in the analysis or writing of this report. 
\subsubsection{\discintname}
The authors have no competing interests to declare that
are relevant to the content of this article.
\end{credits}

\bibliographystyle{splncs04}
\bibliography{mybibliography}
\end{document}